\pdfoutput=1

\documentclass[11pt]{article}

\usepackage{acl}

\usepackage{times}
\usepackage{latexsym}
\usepackage{graphicx}
\usepackage{amsfonts}
\usepackage{amsmath}
\usepackage{multicol}
\usepackage{multirow}
\usepackage{bm}
\usepackage{hyperref}
\usepackage{booktabs}

\usepackage[T1]{fontenc}

\usepackage[utf8]{inputenc}

\usepackage{microtype}

\title{Neural Machine Translation with Phrase-Level Universal Visual Representations}

\author{
    Qingkai Fang\textsuperscript{\rm 1,2},
    Yang Feng\textsuperscript{\rm 1,2}\thanks{ $\;\;$Corresponding author: Yang Feng.} \\
    \textsuperscript{\rm1} Key Laboratory of Intelligent Information Processing \\ Institute of Computing Technology, Chinese Academy of Sciences (ICT/CAS) \\
    \textsuperscript{\rm2} University of Chinese Academy of Sciences, Beijing, China \\
    \texttt{\{fangqingkai21b,fengyang\}@ict.ac.cn} \\
}

\usepackage{lipsum}
\newcommand\blfootnote[1]{%
  \begingroup
  \renewcommand\thefootnote{}\footnote{#1}%
  \addtocounter{footnote}{-1}%
  \endgroup
}

\begin{document}
\maketitle

\blfootnote{\noindent Code is publicly available at \url{https://github.com/ictnlp/PLUVR}.}

\begin{abstract}

Multimodal machine translation (MMT) aims to improve neural machine translation (NMT) with additional visual information, but most existing MMT methods require paired input of source sentence and image, which makes them suffer from shortage of sentence-image pairs. In this paper, we propose a phrase-level retrieval-based method for MMT to get visual information for the source input from existing sentence-image data sets so that MMT can break the limitation of paired sentence-image input. Our method performs retrieval at the phrase level and hence learns visual information from pairs of source phrase and grounded region, which can mitigate data sparsity. Furthermore, our method employs the conditional variational auto-encoder to learn visual representations which can filter redundant visual information and only retain visual information related to the phrase. Experiments show that the proposed method significantly outperforms strong baselines on multiple MMT datasets, especially when the textual context is limited. 


\end{abstract}

\section{Introduction}


Multimodal machine translation (MMT) introduces visual information into neural machine translation (NMT), which assumes that additional visual modality could improve NMT by grounding the language into a visual space \citep{lee2018emergent}. However, most existing MMT methods require additional input of images to provide visual representations, which should match with the source sentence. Unfortunately, in practice it is difficult to get this kind of pairwise input of text and images which hinders the applications of MMT. What is worse, to train an MMT model, the training data still involves the target sentence besides the source sentence and the image, which is costly to collect. As a result, the MMT model is usually trained on a small Multi30K \citep{elliott-etal-2016-multi30k} data set, which limits the performance of MMT. Therefore, it is necessary to utilize the separate image data set to obtain visual representations to break the constraints of pairwise input.

Towards this end, some researchers \citep{Zhang2020Neural, wu-etal-2021-good} propose to integrate a retrieval module into NMT, which retrieve images related to the source sentence from existing sentence-image pairs as complementary input, and then use a pre-trained convolutional neural network (CNN) to encode the images. 
However, such sentence-level retrieval usually suffers from sparsity as it is difficult to get the images that properly match with the source sentence. Besides, visual features outputted by the CNN contain richer information (\emph{e.g.}, color, size, shape, texture, and background) than the source text, thus encoding them in a bundle without any filtering will introduce noise into the model.

To solve these problems, we propose a novel retrieval-based method for MMT to learn phrase-level visual representations for the source sentence, which can mitigate the aforementioned problems of sparse retrieval and redundant visual representations. For the sparsity problem, our method retrieves the image at the phrase level and only refers to the grounded region in the image related with the phrase. For the redundancy problem, our method employs the conditional variational auto-encoder to force the learned representations to properly reconstruct the source phrase so that the learned representations only retain the information related to the source phrase . Experiments on Multi30K \citep{elliott-etal-2016-multi30k} show that the proposed method gains significant improvements over strong baselines. When the textual context is limited, it achieves up to 85\% gain over the text-only baseline on the BLEU score. Further analysis demonstrates that the proposed method can obtain visual information that is more related to translation quality.

\section{Phrase-Guided Visual Representation}
\label{sec:phraseuvr}

We use phrase-level visual representation to improve NMT. In this section, we will introduce our proposed \emph{phrase-guided visual representation}. We first build a phrase-level image set, and then introduce a latent-variable model to learn a phrase-guided visual representation for each image region.

\subsection{Phrase-Level Image Set}
\label{sec:imgset}

Our phrase-level image set is built from the training set of Multi30K, which contains about 29K bilingual sentence-image pairs. We only use the images $\mathbf{e}$ and source descriptions $\mathbf{x}$ from them, which is denoted as $\mathcal{D}=\{(\mathbf{x}_i, \mathbf{e}_i)\}_{i=1}^N$. We extract <noun phrase, image region> pairs from <sentence, image> pairs in $\mathcal{D}$ to build our phrase-level image set, which is denoted as $\mathcal{D}_p$.

\begin{figure}[t]
    \centering
    \includegraphics[width=\linewidth]{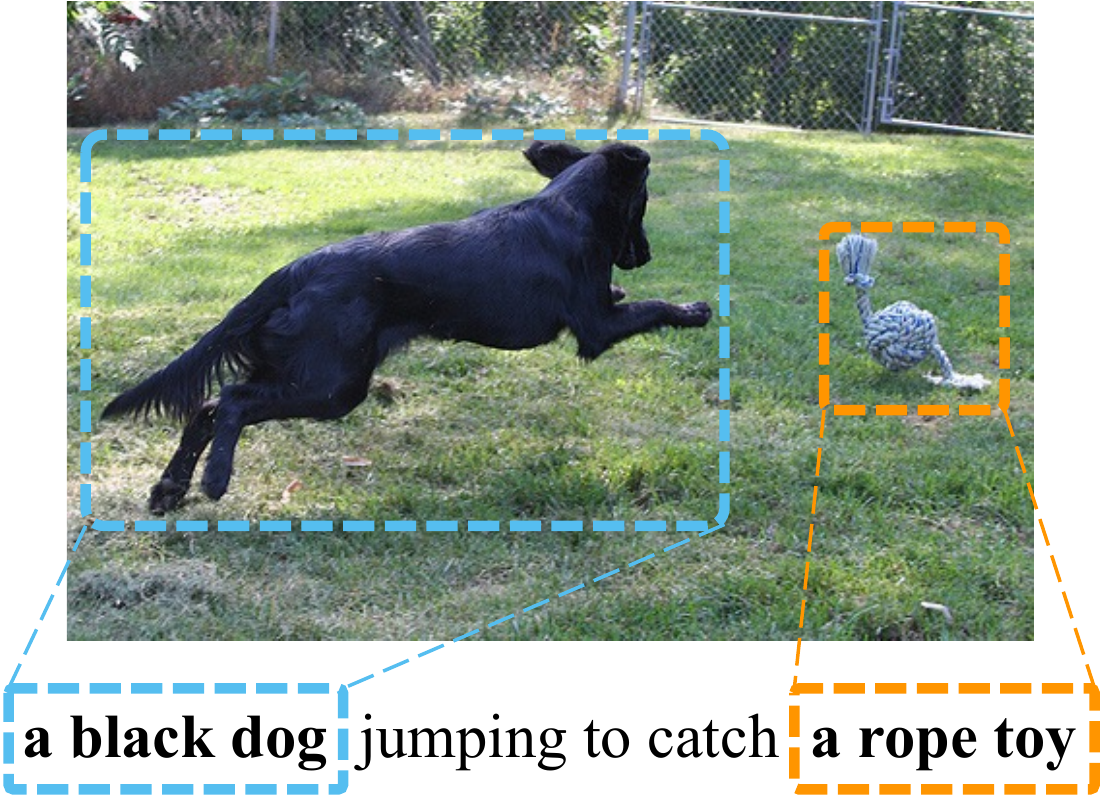}
    \caption{Example of extracting <noun phrase, image region> pairs from existing <sentence, image> pairs.}
    \label{fig:extraction}
\end{figure}

For each sentence $\mathbf{x}_i$, we use an open-source library \emph{spaCy}\footnote{\url{https://spacy.io}} to identify the noun phrases, which is denoted as $\mathbf{P}_i=(\mathbf{p}^i_1, \mathbf{p}^i_2, ..., \mathbf{p}^i_{t_i})$, where $t_i$ is the number of noun phrases in $\mathbf{x}_i$. For each noun phrase $\mathbf{p}^i_j$, we detect the corresponding region $\mathbf{r}^i_j$ from the paired image $\mathbf{e}_i$ using the visual grounding toolkit \citep{yang2019fast}. Then $(\mathbf{p}^i_j, \mathbf{r}^i_j)$ is added to our phrase-level image set $\mathcal{D}_p$. Figure \ref{fig:extraction} illustrates an example.

Finally, we obtain the phrase-level image set $\mathcal{D}_p=\{(\mathbf{p}_i, \mathbf{r}_i)\}_{i=1}^T$, where $T=\sum_{i=1}^Nt_i$. It contains about 102K pairs in total.

\subsection{Latent-Variable Model}
\label{sec:sevr}

For an image region $\mathbf{r}$, we can obtain the visual features $\mathbf{v}$ with a pre-trained ResNet-101 Faster R-CNN \citep{he2016deep, NIPS2015_14bfa6bb}, which contains rich visual information (\emph{e.g.}, color, size, shape, texture, and background). However, we should not pay much attention to the visual information not mentioned in the corresponding phrase, which will introduce too much noise and even be harmful to NMT. Therefore, we further introduce a continuous latent variable to explicitly model the semantic information of image regions under the guidance of phrases. We adopt the framework of conditional variational auto-encoder (CVAE) \citep{Kingma2014, NIPS2015_8d55a249_cvae} to maximize the conditional marginal log-likelihood $\log p(\mathbf{p}|\mathbf{v})=\log \int_z p(\mathbf{p}|z,\mathbf{v})p(z|\mathbf{v})dz$ by maximizing the evidence lowerbound (ELBO):
\begin{equation}
\begin{aligned}
    \mathcal{L}_{cvae}(\omega, \phi, \theta)=&\mathbb{E}_{z\sim q_{\phi}(\mathbf{z}|\mathbf{p},\mathbf{v})}[\log p_{\theta}(\mathbf{p}|\mathbf{z},\mathbf{v})] \\
    &-{\rm KL}[q_{\phi}(\mathbf{z}|\mathbf{p},\mathbf{v})\|p_{\omega}(\mathbf{z}|\mathbf{v})],
\end{aligned}
\end{equation}
where $p_{\omega}(\mathbf{z}|\mathbf{v})$ is the prior, $q_{\phi}(\mathbf{z}|\mathbf{p},\mathbf{v})$ is an approximate posterior and $p_{\theta}(\mathbf{p}|\mathbf{z},\mathbf{v})$ is the decoder. 
The prior $p_{\omega}$ is modeled as a Gaussian distribution:
\begin{gather}
    p_{\omega}(\mathbf{z}|\mathbf{v})=\mathcal{N}(\mathbf{z};\bm{\mu}_p(\mathbf{v}), \bm{\sigma}_p(\mathbf{v})^2\mathbf{I}), \\
    \bm{\mu}_p(\mathbf{v})={\rm Linear}(\mathbf{v}), \\
    \bm{\sigma}_p(\mathbf{v})={\rm Linear}(\mathbf{v}),
\end{gather}
where ${\rm Linear}(\cdot)$ denotes linear transformation. The approximate posterior $q_{\phi}$ is also modeled as a Gaussian distribution:
\begin{gather}
    q_{\phi}(\mathbf{z}|\mathbf{p},\mathbf{v})=\mathcal{N}(\mathbf{z};\bm{\mu}_q(\mathbf{p},\mathbf{v}), \bm{\sigma}_q(\mathbf{p},\mathbf{v})^2\mathbf{I}), \\
    \bm{\mu}_q(\mathbf{p},\mathbf{v})={\rm Linear}([{\rm RNN}(\mathbf{p}),\mathbf{v}]), \\
    \bm{\sigma}_q(\mathbf{p},\mathbf{v})={\rm Linear}([{\rm RNN}(\mathbf{p}),\mathbf{v}]),
\end{gather}
where ${\rm RNN}(\cdot)$ denotes a single-layer unidirectional recurrent neural network (RNN). The final hidden state of RNN is used to compute the mean and variance vectors.

\begin{figure*}[tb]
    \centering
    \includegraphics[width=\textwidth]{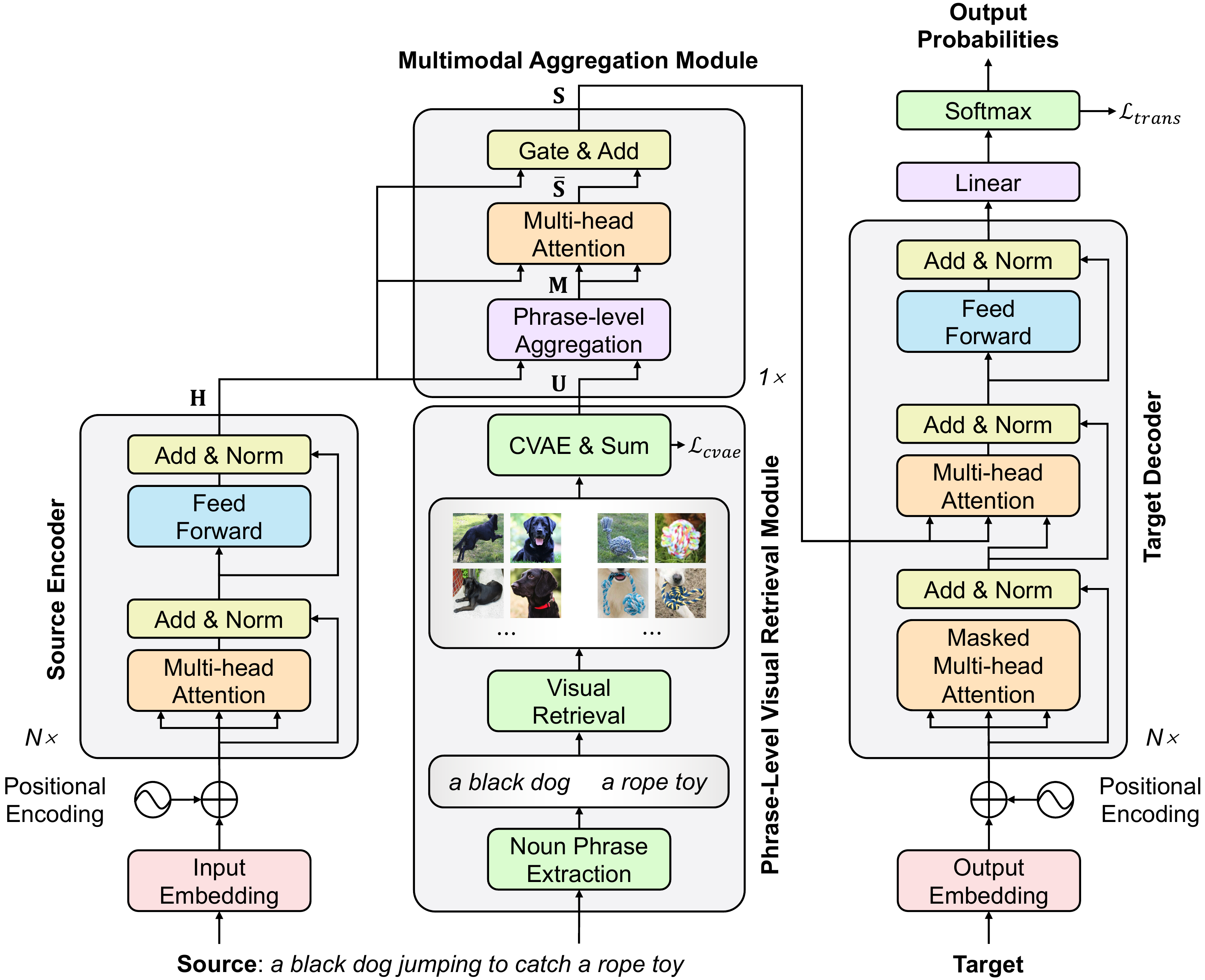}
    \caption{Overview of our proposed method.}
    \label{fig:overview}
\end{figure*}

To be able to update the parameters using backpropagation, we use the reparameterization trick \citep{Kingma2014} to sample $\mathbf{z}$ from $q_{\phi}$:
\begin{equation}
    \mathbf{z}=\bm{\mu}_q+\bm{\sigma}_q\odot\bm{\epsilon}, \bm{\epsilon}\sim\mathcal{N}(0, \mathbf{I}).
\end{equation}

The decoder $p_{\theta}(\mathbf{p}|\mathbf{z},\mathbf{v})$ is also implemented by a single-layer unidirectional RNN. The initial hidden state of decoder RNN is defined as:
\begin{equation}
    \mathbf{s}={\rm Linear}([\mathbf{z}, \mathbf{v}]),
\end{equation}
and then the decoder will reconstruct the phrase $\mathbf{p}$ based on $\mathbf{s}$. 
We refer to $\mathbf{s}$ as \emph{phrase-guided visual representation}, since it pays more attention to the semantic information mentioned in the phrase and filters out irrelevant information. We will describe how to incorporate it into NMT in the next section.

\section{NMT with Phrase-Level Universal Visual Representation}

In this section, we will introduce our retrieval-based MMT method. Specifically, we obtain visual context through our proposed phrase-level visual retrieval, and then learn a universal visual representation for each phrase in the source sentence, which is used to improve NMT. Figure \ref{fig:overview} shows the overview of our proposed method, which is composed of four modules: source encoder, phrase-level visual retrieval module, multimodal aggregation module, and target decoder. The source encoder and target decoder are the same as the encoder and decoder of conventional text-only Transformer \citep{NIPS2017_3f5ee243}. Therefore, we will introduce the phrase-level visual retrieval module and multimodal aggregation module in detail in the rest of this section.

We denote the input source sentence as $\mathbf{x}=(x_1, x_2, ..., x_n)$, the ground truth target sentence as $\mathbf{y}^*=(y_1^*, y_2^*, ..., y_m^*)$ and the generated translation as $\mathbf{y}=(y_1, y_2, ..., y_m)$. The input source sentence $\mathbf{x}$ will be encoded with the source encoder to obtain source sentence representation, which is denoted as $\mathbf{H}=(\mathbf{h}_1, \mathbf{h}_2, ..., \mathbf{h}_n)$.

\subsection{Phrase-Level Visual Retrieval Module}
\label{sec:retrieval}

To obtain the visual context of the source sentence without input paired images, we design a phrase-level visual retrieval module. Specifically, for the input sentence $\mathbf{x}=(x_1, x_2, ..., x_n)$, we identify the noun phrases $\mathbf{\bar{P}}=(\mathbf{\bar{p}}_1, \mathbf{\bar{p}}_2, ..., \mathbf{\bar{p}}_t)$ in $\mathbf{x}$. Each phrase $\mathbf{\bar{p}}_i=(x_{l_i}, x_{l_i+1}, ..., x_{l_i+d_i-1})$ is a continuous list of tokens, where $l_i$ is the index of the first token and $d_i$ is the length of $\mathbf{\bar{p}}_i$. For each noun phrase $\mathbf{\bar{p}}_i$, we will retrieve several relevant <noun phrase, image region> pairs from our phrase-level image set $\mathcal{D}_p$ according to the semantic similarity between phrases, and then use the image regions as visual context. We design a phrase encoder to compute the phrase embedding, which is used to measure the semantic similarity between phrases.

\paragraph{Phrase Encoder}

Our phrase encoder ${\rm Enc_p}(\cdot)$ is based on a pre-trained BERT \citep{devlin-etal-2019-bert}. For a phrase $\mathbf{p}=(p_1, p_2, ..., p_l)$, we first use BERT to encode it into contextual embeddings:
\begin{equation}
    \mathbf{c}_1, \mathbf{c}_2, ..., \mathbf{c}_l = {\rm BERT}(p_1, p_2, ..., p_l),
\end{equation}
then the phrase embedding is the average embedding of all tokens:
\begin{equation}
    {\rm Enc_p}(\mathbf{p})=\frac{1}{l}\sum_{i=1}^l\mathbf{c}_i.
\end{equation}

\paragraph{Visual Retrieval}

For a given phrase $\mathbf{\bar{p}}$, we retrieve top-$K$ relevant <noun phrase, image region> pairs from $\mathcal{D}_p$. For $(\mathbf{p}_i, \mathbf{r}_i)\in\mathcal{D}_p$, the relevance score with given phrase $\mathbf{\bar{p}}$ can be defined as the cosine distance between their phrase embeddings:
\begin{equation}
    {\rm RS}(\mathbf{\bar{p}}, (\mathbf{p}_i, \mathbf{r}_i))=\frac{{\rm Enc_p}(\mathbf{\bar{p}})\cdot{\rm Enc_p}(\mathbf{p}_i)}{\|{\rm Enc_p}(\mathbf{\bar{p}})\|\|{\rm Enc_p}(\mathbf{p}_i)\|},
\end{equation}
then we retrieve top-$K$ relevant pairs for $\mathbf{\bar{p}}$:
\begin{equation}
    \{(\mathbf{p}_{i_k}, \mathbf{r}_{i_k})\}_{k=1}^K 
    =\mathop{{\rm top\text{-}}K}\limits_{i=1..T}({\rm RS}(\mathbf{\bar{p}},(\mathbf{p}_i, \mathbf{r}_i))).
\end{equation}
\paragraph{Universal Visual Representation}For every pair $(\mathbf{p}_{i_k}, \mathbf{r}_{i_k})$, we can obtain the phrase-guided visual representation $\mathbf{s}_{i_k}$ through our latent-variable model as described in Section \ref{sec:sevr}. Finally, the phrase-level \emph{universal visual representation} of $\mathbf{\bar{p}}$ is defined as the weighted sum of all $\{\mathbf{s}_{i_k}\}$:
\begin{equation}
    \mathbf{u}=\frac{1}{K}\sum_{k=1}^{K}{\rm RS}(\mathbf{\bar{p}},(\mathbf{p}_{i_k}, \mathbf{r}_{i_k}))\cdot\mathbf{s}_{i_k}.
\end{equation}
Our universal visual representation considers multi-view visual information from several image regions, which avoids the bias caused by a single image region. Finally, for all phrases $\mathbf{\bar{P}}=(\mathbf{\bar{p}}_1, \mathbf{\bar{p}}_2, ..., \mathbf{\bar{p}}_t)$ in $\mathbf{x}$, we obtain the corresponding universal visual representation $\mathbf{U}=(\mathbf{u}_1, \mathbf{u}_2, ..., \mathbf{u}_t)$.

\subsection{Multimodal Aggregation Module}
\label{sec:model}

Inspired by the recent success of modality fusion in multimodal machine translation \citep{yin-etal-2020-novel, Zhang2020Neural, fang-etal-2022-STEMM}, we design a simple \emph{multimodal aggregation module} to fuse the source sentence representation $\mathbf{H}$ and phrase-level universal visual representation $\mathbf{U}$. At first, we perform a phrase-level aggregation. For each phrase $\mathbf{\bar{p}}_i=(x_{l_i}, x_{l_i+1}, ..., x_{l_i+d_i-1})$, we fuse the universal visual representation $\mathbf{u}_i$ and the textual representation of corresponding tokens $(\mathbf{h}_{l_i}, \mathbf{h}_{l_i+1}, ..., \mathbf{h}_{l_i+d_i-1})$:
\begin{gather}
    \mathbf{m}_i={\rm LayerNorm}(\mathbf{u}_i+\sum_{j=l_i}^{l_i+d_i-1}\mathbf{o}_{ij}\odot\mathbf{h}_j), \\
    \mathbf{o}_{ij}={\rm sigmoid}(\mathbf{W}_1\mathbf{u}_i+\mathbf{W}_2\mathbf{h}_j),
\end{gather}
where $\odot$ denotes element-wise product. Now we obtain the multimodal phrase representation $\mathbf{M}=(\mathbf{m}_1, \mathbf{m}_2, ..., \mathbf{m}_t)$. Afterwards, we apply a multi-head attention mechanism to append $\mathbf{M}$ to the source sentence representation:
\begin{equation}
    \mathbf{\bar{S}}={\rm MultiHead}(\mathbf{H}, \mathbf{M}, \mathbf{M}).
\end{equation}
We then fuse $\mathbf{\bar{S}}$ and $\mathbf{H}$ with a gate mechanism:
\begin{gather}
    \mathbf{S}=\mathbf{H}+\bm{\lambda}\odot\mathbf{\bar{S}}, \\
    \bm{\lambda}={\rm sigmoid}(\mathbf{W}_3\mathbf{H}+\mathbf{W}_4\mathbf{\bar{S}}).
\end{gather}
Finally, $\mathbf{S}$ is fed into our target decoder for predicting the translation. The translation model is trained with a cross-entropy loss:
\begin{equation}
    \mathcal{L}_{trans}=-\sum_{i=1}^m\log p(y_i^*|\mathbf{y}_{<i}, \mathbf{x}).
\end{equation}

\section{Experiments}

\begin{table*}[tb]
\small
\resizebox{\textwidth}{!}{
\begin{tabular}{lllllll}
\toprule
\multicolumn{1}{l}{\multirow{2}{*}{\textbf{Models}}} & \multicolumn{3}{c}{\textbf{EN-DE}} & \multicolumn{3}{c}{\textbf{EN-FR}} \\ 
\multicolumn{1}{l}{}                       & \multicolumn{1}{c}{Test2016} & \multicolumn{1}{c}{Test2017} & \multicolumn{1}{c|}{MSCOCO} & \multicolumn{1}{c}{Test2016} & \multicolumn{1}{c}{Test2017} & \multicolumn{1}{c}{MSCOCO} \\ 
\midrule
\multicolumn{1}{l|}{Transformer \citep{NIPS2017_3f5ee243}} & 39.87 & 31.78 & \multicolumn{1}{l|}{29.36} & 60.51 & 52.44 & 42.49 \\
\multicolumn{1}{l|}{Imagination \citep{elliott-kadar-2017-imagination}} & 39.70\textsuperscript{-0.17} & 32.15\textsuperscript{+0.37} & \multicolumn{1}{l|}{29.76\textsuperscript{+0.40}} & 60.88\textsuperscript{+0.37} & 52.89\textsuperscript{+0.45} & 42.87\textsuperscript{+0.38} \\
\multicolumn{1}{l|}{UVR-NMT \citep{Zhang2020Neural}} & 38.19\textsuperscript{-1.68} & 31.85\textsuperscript{+0.07} & \multicolumn{1}{l|}{28.55\textsuperscript{-0.81}}  & 60.02\textsuperscript{-0.49} & 51.50\textsuperscript{-0.94} & 43.22\textsuperscript{+0.73} \\
\multicolumn{1}{l|}{Ours} & \textbf{40.30\textsuperscript{+0.43}} & \textbf{33.45\textsuperscript{+1.67}**} & \multicolumn{1}{l|}{\textbf{30.28\textsuperscript{+0.92}}}  & \textbf{61.31\textsuperscript{+0.80}*} & \textbf{53.15\textsuperscript{+0.71}*} & \textbf{43.65\textsuperscript{+1.16}*} \\ 
\bottomrule
\end{tabular}}
\centering
\caption{BLEU scores on Multi30K dataset. * and ** mean the improvements over Transformer \cite{NIPS2017_3f5ee243} baseline is statistically
significant ($p < 0.05$ and $p < 0.01$, respectively).}
\label{tab:multi30k-main}
\end{table*}

We conduct experiments on the following datasets:

\paragraph{Multi30K}

Multi30K dataset contains bilingual parallel sentence pairs with image annotations, where each image is paired with one English description and the translations in German and French. Training, validation and test sets contain 29,000, 1,014, and 1,000 instances, respectively. We also report the results on the WMT17 test set and the ambiguous MSCOCO test set, which contain 1,000 and 461 instances respectively. 

\paragraph{WMT16 EN-DE}

WMT16 EN-DE dataset contains about 4.5M sentence pairs. We choose \emph{newstest2013} for validation and \emph{newstest2014} for test. 

\paragraph{WMT16 EN-RO}

WMT16 EN-RO dataset contains about 0.6M sentence pairs. We choose \emph{newsdev2016} for validation and \emph{newstest2016} for test.

For all the above datasets, all sentences are tokenized and segmented into subwords units using byte-pair encoding (BPE) \citep{sennrich-etal-2016-neural}. The vocabulary is shared for source and target languages, with a size of 10K for Multi30K, and 40K for WMT16 EN-DE and WMT16 EN-RO. 

\subsection{System Settings}

\paragraph{Model Implementation}

For the latent-variable model, the image region is encoded with a pre-trained ResNet101 Faster-RCNN \citep{he2016deep, NIPS2015_14bfa6bb}. Both the phrase encoder and decoder are implemented using a single-layer unidirectional RNN with 512 hidden states. The size of the latent variable is set to 64. The batch size is 1024, and the learning rate is 5e-5. We train the model up to 200 epochs with Adam optimizer \citep{DBLP:journals/corr/KingmaB14}. We adopt \emph{KL cost annealing} and \emph{word dropout} tricks to alleviate the posterior collapse problem following \citet{bowman-etal-2016-generating}. The annealing step is set to 20000 and the word dropout is set to 0.1. Note that the phrases are segmented using the same BPE vocabulary as that for each source language.

For the translation model, we use Transformer \citep{NIPS2017_3f5ee243} as our baseline. Both encoder and decoder contain 6 layers. The number of attention heads is set to 4. The dropout is set to 0.3, and the value of label smoothing is set to 0.1. For the visual retrieval module, we retrieve top-5 image regions for each phrase. We use Adam optimizer \citep{DBLP:journals/corr/KingmaB14} to tune the parameters. The learning rate is varied under a warm-up strategy with 2,000 steps. We train the model up to 8,000, 20,000, and 250,000 steps for Multi30K, WMT16 EN-RO, and WMT16 EN-DE, respectively. We average the checkpoints of last 5 epochs for evaluation. We use beam search with a beam size of 4. Different from previous work, we use sacreBLEU\footnote{\url{https://github.com/mjpost/sacrebleu}} \citep{post-2018-call} to compute the BLEU \citep{papineni-etal-2002-bleu} scores and the statistical significance of translation results with paired bootstrap resampling \citep{koehn-2004-statistical} for future standard comparison across papers. Specifically, we measure case-insensitive detokenized BLEU for Multi30K (sacreBLEU signature: nrefs:1 | bs:1000 | seed:12345 | case:lc | eff:no | tok:13a | smooth:exp | version:2.0.0)\footnote{This is because the official pre-processing script of Multi30K dataset lowercases the corpus, see \url{https://github.com/multi30k/dataset/blob/master/scripts/task1-tokenize.sh}} and case-sensitive detokenized BLEU for WMT datasets (sacreBLEU signature: nrefs:1 | bs:1000 | seed:12345 | case:mixed | eff:no | tok:13a | smooth:exp | version:2.0.0).

All models are trained and evaluated using 2 RTX3090 GPUs. We implement the translation model based on \emph{fairseq}\footnote{\url{https://github.com/pytorch/fairseq}} \citep{ott-etal-2019-fairseq}. We train latent-variable model and translation model individually.

\subsection{Baseline Systems}

Our baseline is the text-only Transformer \citep{NIPS2017_3f5ee243}. Besides, we implement \textbf{Imagination} \citep{elliott-kadar-2017-imagination} and \textbf{UVR-NMT} \citep{Zhang2020Neural} based on Transformer, and compare our method with them. The details of these methods can be found in Section \ref{sec:relatedworks}. We use the same configuration for all baseline systems as our model.

\section{Results and Analysis}

\subsection{Results on Multi30K Dataset}

Table \ref{tab:multi30k-main} shows the results on Multi30K. Our proposed method significantly outperforms the Transformer \citep{NIPS2017_3f5ee243} baseline, demonstrating that our proposed phrase-level universal visual representation can be helpful to NMT. Our method also surpass \textbf{Imagination} \citep{elliott-kadar-2017-imagination} and \textbf{UVR-NMT} \citep{Zhang2020Neural}. We consider it is mainly due to the following reasons. First, our phrase-level visual retrieval can obtain strongly correlated image regions instead of weakly correlated whole images. Second, our phrase-level universal visual representation considers visual information from multiple image regions and pays more attention to the semantic information mentioned in the phrases. Last, our phrase-level aggregation module makes it easier for the translation model to exploit the visual information. 

\begin{figure}[t]
    \centering
    \includegraphics[width=\linewidth]{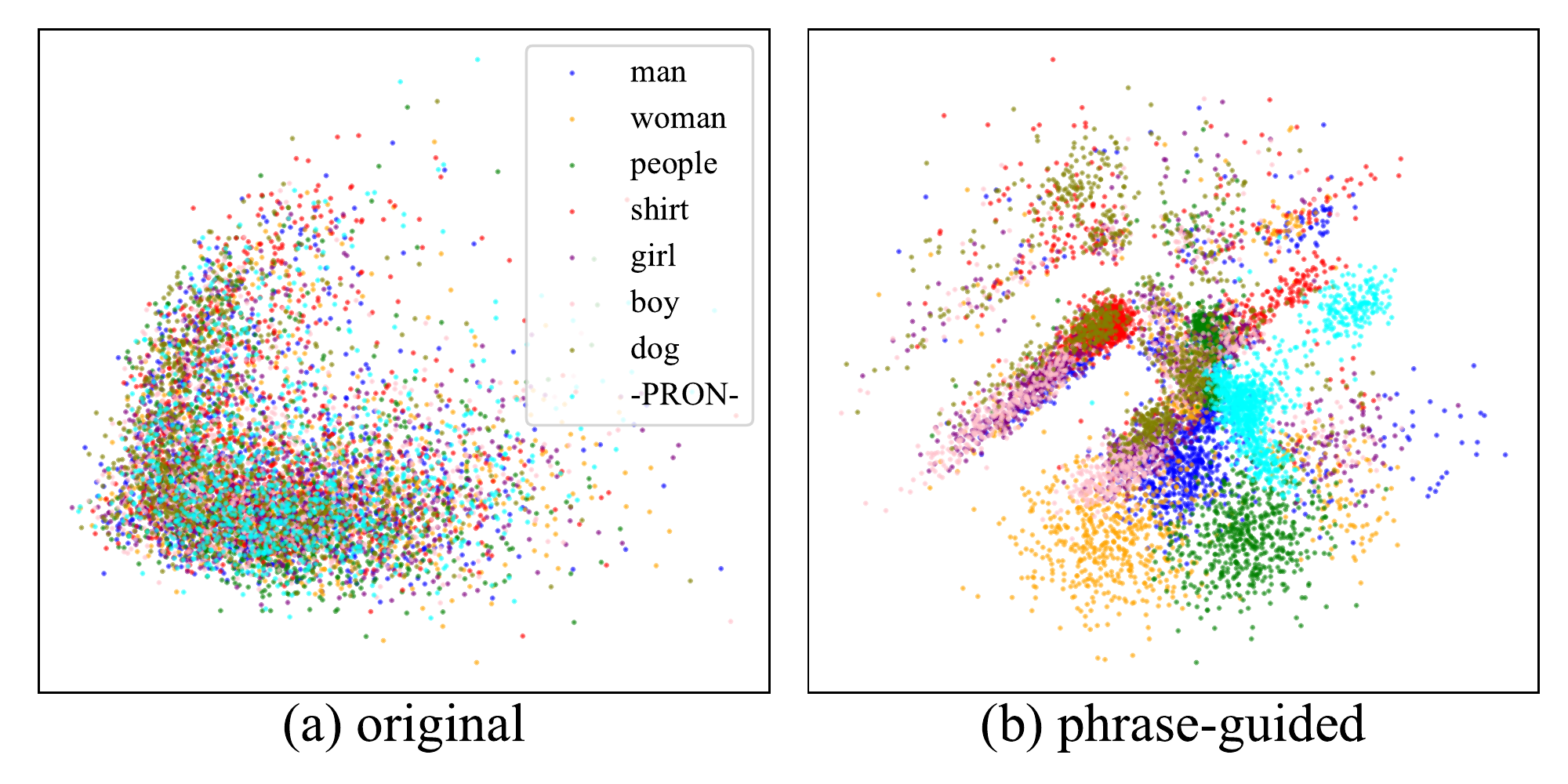}
    \caption{Visualization of different visual representations.}
    \label{fig:pca}
\end{figure}

\subsection{Effects of Latent-Variable Model}

In Section \ref{sec:sevr}, we introduce a latent-variable model to learn a phrase-guided visual representation for each image region. To understand how it improves the model performance compared with original visual features, we visualize the representations by reducing the dimension with Principal Component Analysis (PCA). Specifically, for all <noun phrase, image region> pairs in $\mathcal{D}_p$, we cluster the image regions by the head\footnote{\url{https://en.wikipedia.org/wiki/Head_(linguistics)}} of noun phrases. We select top-8 clusters according to their size, and randomly sample 1000 image regions for each cluster. As shown in Figure \ref{fig:pca}, the original visual features of different clusters are mixed together, indicating that they contains too much irrelevant information. In contrast, our proposed phrase-guided visual representations, which pay more attention to the semantic information, form several clusters according to their heads.

\begin{figure}[t]
    \centering
    \includegraphics[width=\linewidth]{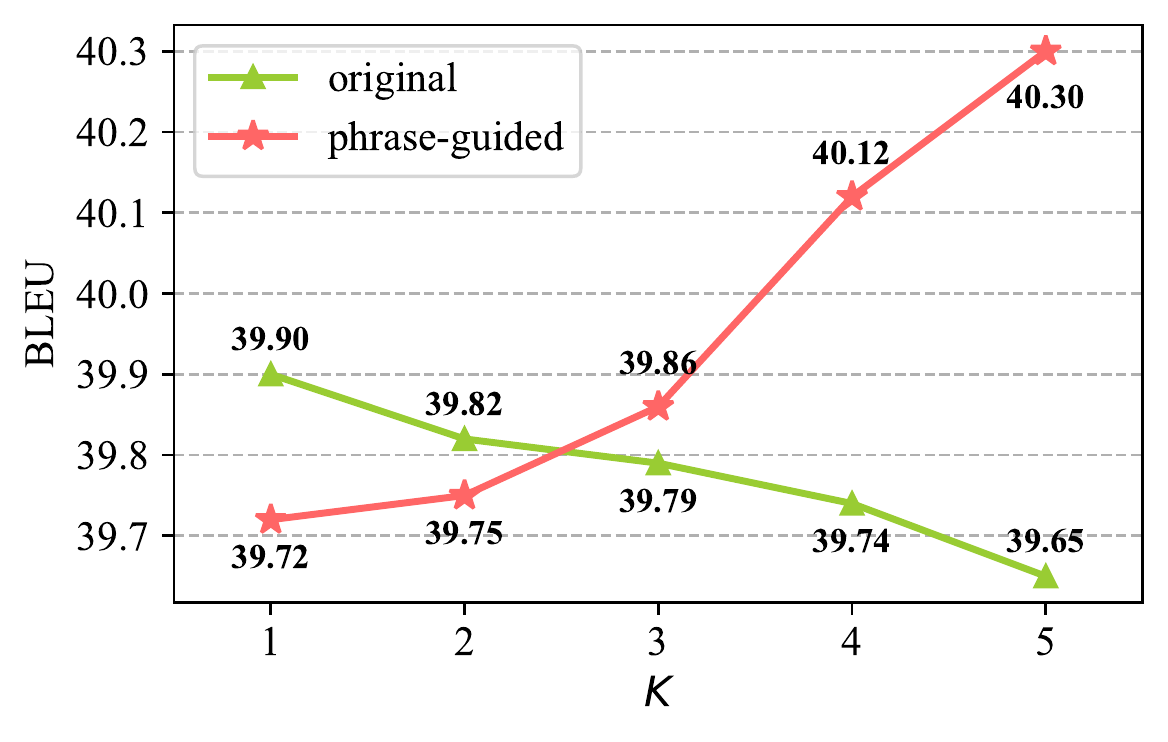}
    \caption{BLEU scores with different number of retrieved image regions $K$. Phrase-guided visual representations achieve better performance as $K$ increases.}
    \label{fig:imgcnt}
\end{figure}

Combined with our visual retrieval module, we found that as the number of retrieved image regions $K$ increases, the BLEU score keeps decreasing when we use original visual features, while increasing when we use our proposed phrase-guided visual representations, which is shown in Figure \ref{fig:imgcnt}. We believe the decrease of BLEU score is due to the irrelevant information in original visual features, and thus directly sum them together will introduce too much noise. Our method filters out those irrelevant information, and multiple image regions could avoid the bias caused by a single one, which leads to the increase of BLEU score. However, we don't observe further improvements when using more image regions.

\begin{figure*}[tb]
    \centering
    \includegraphics[width=\textwidth]{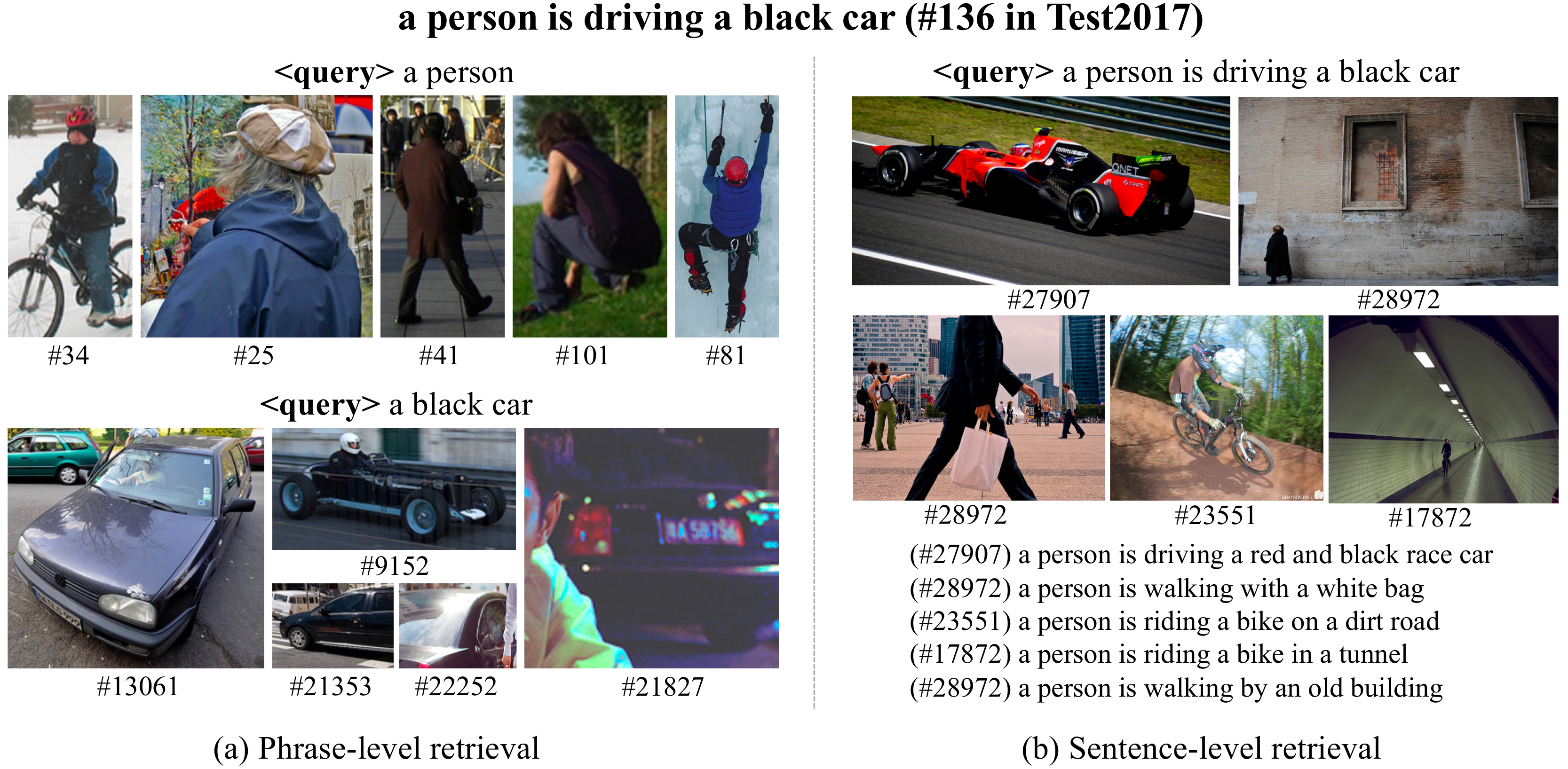}
    \caption{Example of different levels of retrieval. We denote the index of retrieved image (regions) in the training set of Multi30K with \#id.}
    \label{fig:example}
\end{figure*}

\subsection{Source-Degradation Setting}

We further conduct experiments under source-degradation setting, to verify the effectiveness of our method when the source textual context is limited. Following \citet{wu-etal-2021-good}, we mask the visually grounded tokens in the source sentence, which affects around 43\% of tokens in Multi30K. As shown in Table \ref{tab:degrade}, our method achieves almost 85\% improvements over the text-only Transformer baseline. It means our proposed phrase-level universal visual representation can fill in the missing information effectively.

\begin{table}[tb]
\resizebox{\linewidth}{!}{
\begin{tabular}{l|lll}
\toprule
Models       & \multicolumn{1}{c}{Test2016} & \multicolumn{1}{c}{Test2017} & \multicolumn{1}{c}{MSCOCO}  \\ 
\midrule
Transformer  & 10.42    & 8.59  & 7.08              \\
Ours        & \textbf{19.41\textsuperscript{+8.99}}    & \textbf{13.67\textsuperscript{+5.08}} & \textbf{12.23\textsuperscript{+5.15}}                       \\ 
\bottomrule
\end{tabular}}
\centering
\caption{BLEU scores on Multi30K En-De under source-degradation setting.}
\label{tab:degrade}
\end{table}

\subsection{Phrase-Level vs. Sentence-Level Retrieval}
\label{sec:phrase-vs-sentence}

To prove the effectiveness of phrase-level retrieval, we implement a sentence-level variant of our method. In this variant, we switch the latent-variable model, retrieval module and aggregation module from phrase-level to sentence-level. In this way, we retrieve several images as visual context to help the translation. As shown in Table \ref{tab:sentence-variant}, the sentence-level variant \emph{Ours-sentence} performs worse than \emph{Ours}, especially in the case of source-degradation setting. We believe it is because phrase-level retrieval can obtain more relevant image regions as visual context, which contain less noise and can be integrated into textual representations more precisely. In contrast, sentence-level retrieval leads to images with much irrelevant information, and makes it difficult for the model to capture the fine-grained semantic correspondences between images and descriptions. To understand this difference more intuitively, we give an example in Figure \ref{fig:example}. As we can see, for the input sentence, phrase-level retrieval can obtain closely related image regions for noun phrases \emph{a person} and \emph{a black car}, while the results of sentence-level retrieval are actually weakly related with the input sentence.

\begin{table}[tb]
\begin{tabular}{l|ll}
\toprule
Models       & \multicolumn{1}{c}{Test2016} & \multicolumn{1}{c}{Test2016 (Mask)} \\ 
\midrule
Transformer  & 39.87    & 10.42    \\
Ours-sentence        & 40.02\textsuperscript{+0.15}   & 11.52\textsuperscript{+1.10}*  \\
Ours        & \textbf{40.30\textsuperscript{+0.43}}   & \textbf{19.41\textsuperscript{+8.99}**}     \\ 
\bottomrule
\end{tabular}
\centering
\caption{BLEU scores on Multi30K En-De Test2016. (Mask) indicates source-degradation setting. * and ** mean the improvements over Transformer \cite{NIPS2017_3f5ee243} baseline is statistically
significant ($p < 0.05$ and $p < 0.01$, respectively).}
\label{tab:sentence-variant}
\end{table}

\subsection{Results on WMT News Datasets}

\begin{figure}[t]
    \centering
    \includegraphics[width=\linewidth]{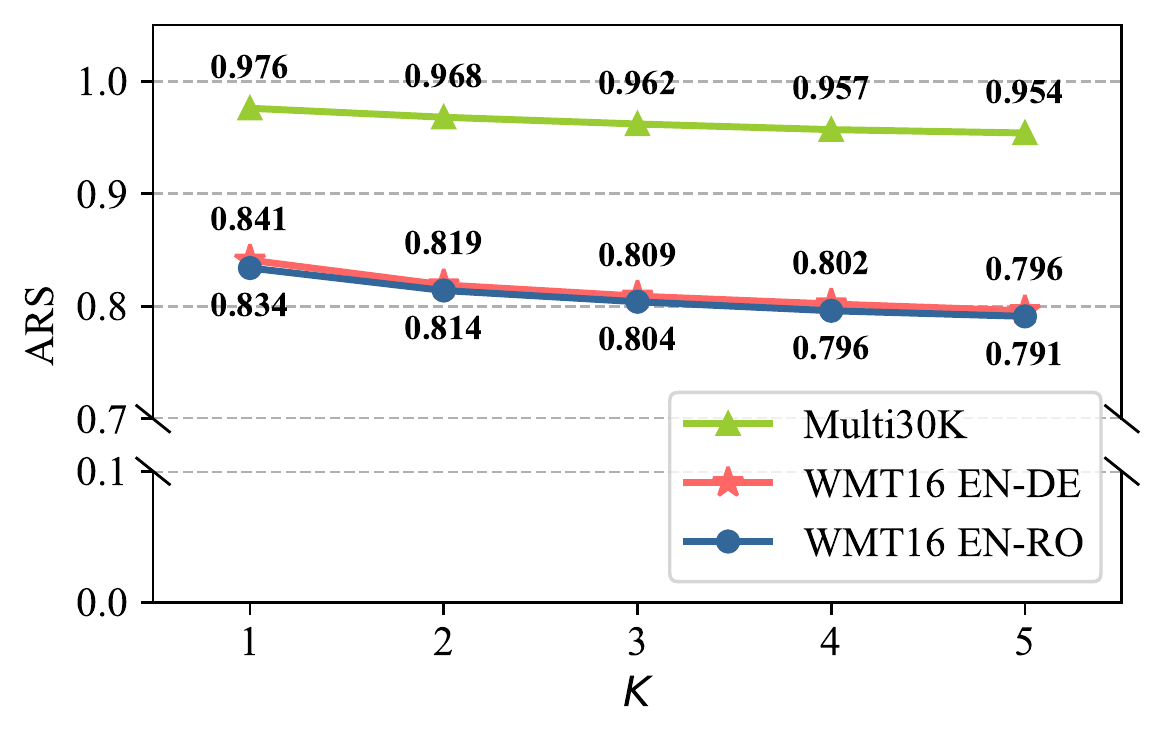}
    \caption{Average relevance scores (ARS) during visual retrieval for all phrases in the validation set.}
    \label{fig:rs}
\end{figure}

Finally, we conduct experiments on WMT16 EN-DE and WMT16 EN-RO datasets. As shown in Table \ref{tab:wmt-main}, we observe that both \citet{Zhang2020Neural} and our method only achieve marginal improvements compared with text-only Transformer baseline. We consider that there are two main reasons. On the one hand, most of tokens in such news text are not naturally related to specific visual contents. We found that the percentage of visual grounded tokens in the training set of WMT16 EN-DE is only 7\% (vs. 43\% in Multi30K), so the contribution of visual information is indeed limited. On the other hand, the news text is far from the descriptive text in Multi30K. In this way, the retrieved image regions are actually weakly correlated with the source phrase. We did some analysis to verify our hypotheses. As described in Section \ref{sec:retrieval}, we retrieve top-$K$ pairs for each phrase according to the relevance scores. We define the average relevance scores (ARS) as follows:
\begin{equation}
    \text{ARS}(k) = \mathbb{E}_{\mathbf{p}\in\mathcal{D}_{\rm val}} {\rm RS}(\mathbf{p}, (\mathbf{p}_{i_k}, \mathbf{r}_{i_k})),
\end{equation}
which means the average relevance scores for all phrases in the validation set. As shown in Figure \ref{fig:rs}, ARS on WMT news datasets are much lower than that on Multi30K, which proves that the gap between news text and descriptive text does exists. 

\begin{table}[tb]
\begin{tabular}{l|ll}
\toprule
Models       & \multicolumn{1}{c}{EN-DE} & \multicolumn{1}{c}{EN-RO} \\ 
\midrule
Transformer  & 26.54                      & 32.67                     \\
UVR-NMT & 26.89\textsuperscript{+0.35}                    & 32.93\textsuperscript{+0.26}                     \\
Ours        & \textbf{26.97\textsuperscript{+0.43}}                      & \textbf{33.18\textsuperscript{+0.51}}                     \\ 
\bottomrule
\end{tabular}
\centering
\caption{BLEU scores on WMT16 EN-DE and WMT16 EN-RO dataset.}
\label{tab:wmt-main}
\end{table}


\section{Related Work}
\label{sec:relatedworks}

Multimodal machine translation (MMT) aims to enhance NMT \citep{NIPS2017_3f5ee243, zhang-etal-2019-bridging, li-etal-2021-mixup-decoding} with additional visual context. Since the release of Multi30K \citep{elliott-etal-2016-multi30k} dataset, researchers have proposed many MMT methods. Early methods \citep{huang-etal-2016-attention, calixto-liu-2017-incorporating, caglayan2016multimodal, calixto-etal-2016-dcu, caglayan-etal-2017-lium, libovicky-helcl-2017-attention, DBLP:journals/corr/DelbrouckD17, delbrouck-dupont-2017-empirical, zhou-etal-2018-visual, calixto-etal-2017-doubly, helcl-etal-2018-cuni, caglayan-etal-2018-lium} are mainly based on the RNN-based encoder-decoder architecture with attention \citep{bahdanau2014neural}. Recent methods based on Transformer \citep{NIPS2017_3f5ee243} achieve better performance. \citet{yao-wan-2020-multimodal, yin-etal-2020-novel, liu2021gumbel} design multimodal encoder to fuse the textual and visual information during encoding. \citet{ive-etal-2019-distilling, 10.1145/3394171.3413715} enhance the decoder with deliberation networks \citep{NIPS2017_c6036a69} or capsule networks \citep{NIPS2017_2cad8fa4_capsule} to better utilize visual information during decoding.  \citet{caglayan-etal-2021-cross} propose a cross-lingual visual pre-training method and fine-tuned for MMT. 
It is worth noting that some of previous works \citep{ive-etal-2019-distilling, 10.1145/3394171.3413715, yin-etal-2020-novel, DBLP:journals/corr/abs-2101-05208, nishihara-etal-2020-supervised, zhao2021neural} adopt regional visual information like us, which shows effectiveness compared with global visual features. The major difference between our method and theirs is that our method is a retrieval-based method, which breaks the reliance on bilingual sentence-image pairs, Therefore, our method is still applicable when the input is text only (without paired images), which is unfortunately not available with those previous methods.

In addition to focusing on model design, \citet{Yang_Chen_Zhang_Sun_2020, nishihara-etal-2020-supervised, DBLP:journals/corr/abs-2101-05208} propose auxiliary loss to allow the model to make better use of visual information. \citet{caglayan-etal-2019-probing, wu-etal-2021-good} conduct systematic analysis to probe the contribution of visual modality. \citet{caglayan-etal-2020-simultaneous, ive-etal-2021-exploiting} focus on improving simultaneous machine translation with visual context.

All of the above methods require a specific image as input to provide visual context, which heavily restricts their applicability. To break this bottleneck, \citet{hitschler-etal-2016-multimodal} propose target-side image retrieval to help the translation. \citet{elliott-kadar-2017-imagination} propose a multitask learning framework \textbf{Imagination} to decomposes the multimodal translation into learning translation and learning visually grounded representation. \citet{calixto-etal-2019-latent} introduce a latent variable and estimate a joint distribution over translations and images. \citet{long2020generative} predict the translation with visual representation generated by a generative adversarial network (GAN) \citep{NIPS2014_5ca3e9b1_gan}. The most closely related work to our method is \textbf{UVR-NMT} \citep{Zhang2020Neural}, which breaks the reliance on bilingual sentence-image pairs. Like some retrieval-enhanced MT \citep{feng-etal-2017-memory, DBLP:journals/corr/GuWCL17} methods, they build a topic-image lookup table from Multi30K, and then retrieve images related to the source sentence as visual context based on the topic words. The central differences between \citet{Zhang2020Neural} and our method are as follows:

\begin{itemize}
    \item First, their method depends on the weak correlation between words and images, which leads to much noise in the retrieved images, while our approach relies on the strong correlation between noun phrases and image regions.
    \item Second, our phrase-level retrieval can obtain more related visual context than their sentence-level retrieval (Section \ref{sec:phrase-vs-sentence}).
    \item Last, their method directly uses visual features extracted by ResNet \citep{he2016deep}, which may introduce too much noise. We adopt a latent-variable model to filter out irrelevant information and obtain a better representation.
\end{itemize}

\section{Conclusion}

In this paper, we propose a retrieval-based MMT method, which learns a phrase-level universal visual representation to improve NMT. Our method not only outperforms the baseline systems and most existing MMT systems, but also breaks the restrictions on input that hinder the development of MMT in recent years. Experiments and analysis demonstrate the effectiveness of our proposed method. In the future, we will explore how to apply our method to other tasks.

\section*{Acknowledgements}
We thank all the anonymous reviewers for their insightful and valuable comments. This work was supported by National Key R\&D Program of China (NO. 2017YFE0192900).

\bibliography{anthology,custom}
\bibliographystyle{acl_natbib}

\end{document}